%% file: elsarticle-template-num.tex
\documentclass[preprint,12pt]{elsarticle}

\usepackage{amssymb}
\usepackage{booktabs}
\usepackage{xcolor}
\usepackage{multirow}
\usepackage{pifont}
\usepackage{diagbox}
\usepackage{amsmath}
\usepackage{adjustbox}
\usepackage{subfig}
\usepackage{hyperref}

\journal{Neural Networks}

\begin{document}

\begin{frontmatter}



\title{ADFQ-ViT: Activation-Distribution-Friendly Post-Training Quantization for Vision Transformers}


\author[nk,kl]{Yanfeng Jiang \fnref{co-first}}
\author[zj]{Ning Sun \fnref{co-first}}
\author[hh]{Xueshuo Xie}
\author[zj]{Fei Yang \corref{cor2}}
\author[nk,hh]{Tao Li \corref{cor1}}

\affiliation[nk]{organization={College of Computer Science, Nankai University},
            city={Tianjin},
            country={China}}

\affiliation[kl]{organization={Tianjin Key Laboratory of Network and Data Security Technology},
            city={Tianjin},
            country={China}}

\affiliation[zj]{organization={Zhejiang Lab},
            city={Hangzhou},
            state={Zhejiang},
            country={China}}

\affiliation[hh]{organization={Haihe Lab of ITAI},
            city={Tianjin},
            country={China}}

\fntext[co-first]{These authors contributed equally to this work}
\cortext[cor1]{Corresponding author at College of Computer Science, Nankai University, Tianjin, China; Email: litao@nankai.edu.cn}
\cortext[cor2]{Corresponding author at Zhejiang Lab, Hangzhou, Zhejiang, China; Email: yangf@zhejianglab.com}

\begin{abstract}
Vision Transformers (ViTs) have exhibited exceptional performance across diverse computer vision tasks, while their substantial parameter size incurs significantly increased memory and computational demands, impeding effective inference on resource-constrained devices. Quantization has emerged as a promising solution to mitigate these challenges, yet existing methods still suffer from significant accuracy loss at low-bit. We attribute this issue to the distinctive distributions of post-LayerNorm and post-GELU activations within ViTs, rendering conventional hardware-friendly quantizers ineffective, particularly in low-bit scenarios. To address this issue, we propose a novel framework called Activation-Distribution-Friendly post-training Quantization for Vision Transformers, ADFQ-ViT. Concretely, we introduce the Per-Patch Outlier-aware Quantizer to tackle irregular outliers in post-LayerNorm activations. This quantizer refines the granularity of the uniform quantizer to a per-patch level while retaining a minimal subset of values exceeding a threshold at full-precision. To handle the non-uniform distributions of post-GELU activations between positive and negative regions, we design the Shift-Log2 Quantizer, which shifts all elements to the positive region and then applies log2 quantization. Moreover, we present the Attention-score enhanced Module-wise Optimization which adjusts the parameters of each quantizer by reconstructing errors to further mitigate quantization error. Extensive experiments demonstrate ADFQ-ViT provides significant improvements over various baselines in image classification, object detection, and instance segmentation tasks at 4-bit. Specifically, when quantizing the ViT-B model to 4-bit, we achieve a 10.23\% improvement in Top-1 accuracy on the ImageNet dataset.
\end{abstract}




\end{frontmatter}


\input{tex/introduction}
\input{tex/related_work}
\input{tex/methodology}
\input{tex/experiments}
\input{tex/conclusion}

\section*{Declaration of competing interest}
The authors declare that they have no known competing financial interests or personal relationships that could have appeared to influence the work reported in this paper.

\section*{Data availability}
Code and data will be made available on request.

\section*{Acknowledgments}
This work is supported by the National Natural Science Foundation of China [grant numbers 62272248, No.U22A6001]; the Shanghai Artificial Intelligence Laboratory [grant numbers No.P22KN00581]; the Natural Science Foundation of Tianjin [grant numbers 23JCQNJC00010].


\bibliographystyle{elsarticle-harv} 
\bibliography{references}





\end{document}

%% file: tex/introduction.tex
\section{Introduction}
Vision Transformers (ViTs) \citep{vit} have achieved outstanding performance across various computer vision tasks such as image classification, object detection, and instance segmentation. Despite this, in contrast to previously common visual backbones like Convolutional Neural Networks (CNNs) \citep{krizhevsky2012imagenet}, ViTs require significantly more memory and computational requirements due to their large number of parameters. Consequently, deploying ViTs in resource-constrained environments or under the constraints of real-time inference remains a challenge.

Model quantization \citep{krishnamoorthi2018quantizing} effectively mitigates memory and computational requirements for inference by transmuting model weights and activations from full-precision to lower-bit representations, wherein it is typically bifurcated into Quantization-Aware Training (QAT) \citep{jacob2018quantization} and Post-Training Quantization (PTQ) \citep{hubara2021accurate}. Although QAT typically yields superior accuracy, it necessitates retraining the model on the entire original training dataset, making it highly resource-intensive. In contrast, PTQ requires only a small subset of unlabeled training data for calibration to quantize a pre-trained model effectively. 

Current the ViTs quantization methods \citep{liu2021post, RepQ-ViT} mainly focus on PTQ; however, existing methods still experience substantial accuracy degradation at low-bit, especially at 4-bit. We have identified that the accuracy degradation at low-bit is predominantly attributable to the distribution of activations in ViTs. Specifically, post-LayerNorm activations exhibit irregular, sporadic outliers, while post-GELU activations demonstrate a non-uniform distribution of positive and negative values. In the context of quantization, hardware-friendly quantizers are often favored for their practical deployment efficiency. Nevertheless, these quantizers impose stringent constraints on data distribution. When the data distribution fails to align with the requirements of quantizers, significant quantization errors ensue, exacerbated by the constrained representation capacity at low-bit. The unfriendly distribution characteristics of activations in ViTs present substantial challenges for employing these quantizers at low-bit, ultimately leading to accuracy loss.

In this paper, we propose a novel Activation-Distribution-Friendly post-training Quantization framework called ADFQ-ViT to address the aforementioned activation distribution challenges in ViTs quantization, aiming to enhance the usability of low-bit. We introduce a Per-Patch Outlier-aware Quantizer tailored for post-LayerNorm activations, effectively reducing quantization error by refining the quantization granularity and preserving outliers in their original full-precision. To tackle the challenge of non-uniform distribution of positive and negative values in post-GELU activations, we design the innovative Shift-Log2 Quantizer, which shifts all elements to positive region, applies the log2 quantizer for quantization, and then shifts them back. Additionally, Attention-score enhanced Module-wise Optimization is presented to further minimize quantization error, optimizing the parameters of weight and activation quantizers to minimize attention score and output error disparities before and after quantization across each module. By integrating these three components, our ADFQ-ViT framework effectively mitigates the quantization error stemming from the distinctive activation distributions in ViTs, thereby enhancing the viability of 4-bit quantization. Extensive experimental results demonstrate that our ADFQ-ViT significantly improves accuracy over all baselines across various ViT, DeiT, and Swin Transformer after quantization on image classification, object detection, and instance segmentation tasks. Notably, under 4-bit quantization setting for image classification on the ImageNet dataset, our method achieves a maximum accuracy improvement of 10.23\% and an average improvement of 3.82\% across seven different models compared to the baseline, highlighting the effectiveness of ADFQ-ViT in low-bit scenarios. Moreover, under 6-bit quantization, the accuracy of ADFQ-ViT after quantization closely approaches that of the original full-precision model.

Our main contributions are summarized as follows:
\begin{itemize}
    \item We design the Per-Patch Outlier-aware Quantizer and the Shift-Log2 Quantizer, effectively addressing the challenges of outliers and irregular distributions in post-LayerNorm activations, as well as the non-uniform distribution of positive and negative values in post-GELU activations.
    \item We introduce the Attention-score enhanced Module-wise Optimization, which optimizes the parameters of weight and activation quantizer to further reduce errors before and after quantization.
    \item We propose a novel Activation-Distribution-Friendly post-training Quantization framework, ADFQ-ViT, which enables efficient quantization of ViTs under low-bit by effectively combining the above three methods.
    \item Experimental results demonstrate that ADFQ-ViT effectively quantizes various ViT variants, achieving improved accuracy compared to all baselines across multiple computer vision tasks after quantization. Particularly, significant improvements in accuracy are observed on image classification task under the 4-bit quantization.
\end{itemize}

%% file: tex/related_work.tex
\section{Related work}
\subsection{Vision Transformers}
ViTs have been widely applied in numerous computer vision tasks due to their impressive performance. Initially, ViT \citep{vit} leverages the Transformer \citep{vaswani2017attention} model to completely replace the CNNs \citep{he2016deep, sarigul2019differential} commonly used in the vision field, achieving superior performance through pre-training on large-scale datasets. DeiT \citep{deit} addresses the substantial data requirements of earlier ViTs by introducing a teacher-student distillation \citep{hinton2015distilling} training strategy, enabling similar performance with training solely on ImageNet \citep{krizhevsky2012imagenet} dataset. Swin Transformer \citep{swin} tackles the high computational demands of ViT for high-resolution images with its shifted window mechanism. In addition to advancements in backbone networks, DETR \citep{carion2020end} utilizes ViT for object detection, creating an end-to-end detection network. Segment Anything Model (SAM) \citep{kirillov2023segment} establishes a foundational model for image segmentation by pre-training on the ultra-large-scale SA-1B dataset. CLIP \citep{radford2021learning} achieves an effective multimodal model by combining visual and language transformers, trained through contrastive learning. Despite the great success of ViTs in computer vision, its parameter size leads to huge computational and memory requirements, making it still challenging to use in real-time and resource-constrained scenarios \citep{papa2024survey}.

\subsection{Quantization for Vision Transformers}
We primarily focus on PTQ \citep{nagel2020up, BRECQ, Qdrop, dettmers2022gpt3, pdquant} in this paper, as it is more widely applicable compared to QAT \citep{qat, lee2021qttnet, kirtas2022quantization} and is predominantly used in current ViTs quantization. PTQforViT \citep{liu2021post} utilizes ranking loss to maintain the order of self-attention pre- and post-quantization. Additionally, it adopts a mixed-precision approach, ultimately achieving accurate quantization under 8-bit. FQ-ViT \citep{lin2021fq} employs a power-of-two factor to address inter-channel variation in pre-LayerNorm and uses Log-Int-Softmax to tackle the extreme non-uniform distribution problem in the attention map, thereby achieving fully 8-bit quantized ViTs. PTQ4ViT \citep{PTQ4ViT} introduces twin uniform quantization for post-Softmax and post-GELU activations, dividing the activations into two distinct quantization ranges, each controlled by a unique scaling factor, thus achieving near-lossless 8-bit quantization. APQ-ViT \citep{ding2022towards} incorporates Bottom-elimination Blockwise Calibration and Matthew-effect Preserving Quantization to reduce quantization errors, significantly improving 6-bit quantization performance. NoisyQuant \citep{liu2023noisyquant} addresses the quantization issues of long-tail activation distributions by adding a fixed uniform noise, further enhancing 6-bit quantization accuracy. RepQ-ViT \citep{RepQ-ViT} improves 4-bit quantization accuracy by separating the quantization and deployment processes through scale reparameterization. During deployment, it converts per-channel quantizer and log$\sqrt{2}$ quantizer into per-tensor and log2 quantizer, respectively. However, current ViTs quantization methods still suffer significant accuracy loss at 4-bit. For instance, RepQ-ViT experiences accuracy loss exceeding 15\% on certain models. The pursuit of 4-bit PTQ for ViTs poses a challenging yet immensely valuable goal, as lower-bit quantization promises substantial performance gains for deployment scenarios.

%% file: tex/methodology.tex
\begin{figure*}[t]
    \centering
    \includegraphics[width=\textwidth]{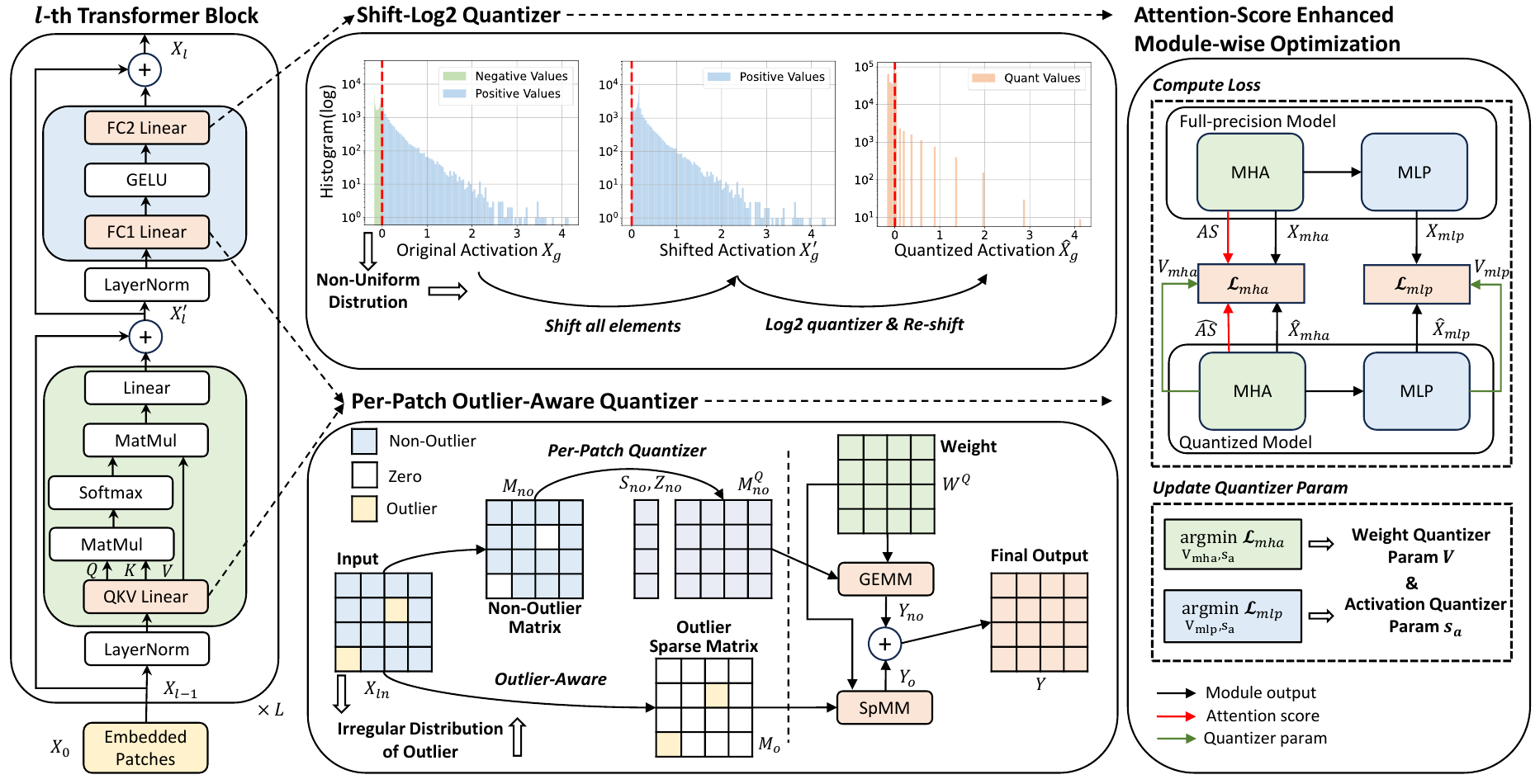}
    \caption{Overview of our ADFQ-ViT framework. We apply Per-Patch Outlier-aware Quantizer for post-LayerNorm activations (QKV and FC1 Linear), Shift-Log2 Quantizer for post-GELU activations (FC2 Linear), and Attention-score enhanced Module-wise Optimization for optimizing the parameters of weight and input activation quantizers.}
    \label{fig:overview}
\end{figure*}

\section{Methodology}
In this section, we comprehensively describe our ADFQ-ViT framework, with its overview illustrated in Figure \ref{fig:overview}. ADFQ-ViT comprises three pivotal components: the Per-Patch Outlier-aware Quantizer, the Shift-Log2 Quantizer, and the Attention-score enhanced Module-wise Optimization. Specifically, Section \ref{subsec:per-patch} introduces the Per-Patch Outlier-aware Quantizer, which addresses the challenge of irregular distributions in post-LayerNorm activations within ViTs quantization. In Section \ref{subsec:shift-log2}, we describe the Shift-Log2 Quantizer, which effectively tackles the problem of the non-uniform distribution of negative and positive values in post-GELU activations. The Attention-score enhanced Module-wise Optimization aims to reduce quantization error by optimizing the parameters of quantizers, as presented in Section \ref{subsec:attention-score}. Lastly, we detail the complete quantization procedure of ADFQ-ViT in Section \ref{subsec:procedure}.

\subsection{Preliminaries}
\label{subsec:preliminaries}
\textbf{Architecture of ViT.} The network architecture of ViT \citep{vit} mainly consists of $L$ cascaded transformer blocks \citep{vaswani2017attention}. Given an input image $I \in \mathbb{R}^{h\times w\times c}$, where $(h, w)$ represents the resolution of the image and $c$ denotes the number of channels, it is necessary to preprocess the image into a sequential form before feeding $I$ into the transformer blocks. The specific approach is to split $I$ into $n$ patches $p \in \mathbb{R}^{h_p\times w_p \times c}$, where $(h_p, w_p)$ represents the resolution of each patch. These patches are then linearly projected into $d$-dimensional vectors, and positional embeddings are added to obtain the final embedded patches. Ultimately, these embedded patches constitute the input $X_0 \in \mathbb{R}^{n \times d}$ to the subsequent transformer blocks (for simplicity, we omit the class embedding in this description). 

Inside each transformer block, there exists a Multi-Head Attention (MHA) module and a Multi-Layer Perceptron (MLP) module. For the $l$-th transformer block, the MHA module incorporates $H$ attention heads. The computational procedure for each head $i$ unfolds as follows:
\begin{equation}
[Q_i, K_i, V_i] = \operatorname{LayerNorm}(X_{l-1})W^{i}_{qkv} + b^{i}_{qkv}
\enskip,
\end{equation}
\begin{equation}
AS_i = \operatorname{Softmax}\left(\frac{Q_iK^T_i}{\sqrt{d_{head}}}\right)V_i
\enskip,
\end{equation}
where $W^{i}_{qkv}$ and $b^{i}_{qkv}$ denote the weights and biases of the QKV Linear layer, $d_{head}=\frac{d}{H}$, and $AS_i$ represents the attention score for head $i$. The output of MHA module is represented as:
\begin{equation}
X_{mha} = \operatorname{concat}([AS_i,...,AS_H])W_o + b_o
\enskip,
\end{equation}
where $W_o$ and $b_o$ denote the weights and bias of the output projection layer in the MHA module.

Within the MLP module, two Linear layers, namely FC1 and FC2, are present. The specific computational process is as follows:
\begin{equation}
X'_l = \operatorname{LayerNorm}(X_{mha}+X_{l-1})
\enskip,
\end{equation}
\begin{equation}
X_{fc1} = \operatorname{GELU}(X'_lW_{fc1} + b_{fc1})
\enskip,
\end{equation}
\begin{equation}
X_{fc2}=X_{fc1}W_{fc2}+b_{fc2}
\enskip,
\end{equation}
where $W_{fc1}$ and $b_{fc1}$ represent the weights and bias of the FC1 Linear layer, while $W_{fc2}$ and $b_{fc2}$ denote the weights and bias of the FC2 Linear layer. The final output of the $l$-th transformer block is given by:
\begin{equation}
X_l = X'_l+X_{fc2}
\enskip.
\end{equation}

\textbf{Quantizer for PTQ.}
PTQ \citep{nagel2021white} primarily involves using a quantizer $\mathcal{Q}$ to convert the original floating-point values $X_f$ into $k$-bit fixed-point integers $X_{Q}$:
\begin{equation}
X_{Q} = \mathcal{Q}(X_f|k)
\enskip.
\end{equation}
Different quantizers $\mathcal{Q}$ adopt distinct mapping strategies, exerting a notable influence on both accuracy and performance. Consequently, it is crucial to select an appropriate $\mathcal{Q}$ based on the data distribution and practical application scenarios. Considering the activation distribution characteristics of ViTs and actual deployment performance, previous ViTs quantization methods typically employ uniform quantizer and log2 quantizer.

The uniform quantizer $\mathcal{UQ}$ is particularly well-suited for scenarios where the data distribution is relatively uniform and benefits from highly efficient hardware implementations. The computation process is as follows:
\begin{equation}
X_{Q} = \mathcal{UQ}(X_f|k) = \operatorname{clamp}(0,2^{k}-1,\lfloor \frac{X_f}{s} \rceil + z)
\enskip,
\end{equation}
\begin{equation}
s = \frac{\operatorname{max}(X_f)-\operatorname{min}(X_f)}{2^k - 1}, z = \lfloor \frac{-\operatorname{min}(X_f)}{s} \rceil
\enskip,
\end{equation}
\begin{equation}
\operatorname{clamp}(l,u,x) = \begin{cases}
    l, & x \leq l \\
    x, & l \leq x \leq u \\
    u, & x \geq u
\end{cases}
\enskip,
\end{equation}
where $s$ is the scaling factor, $z$ is the zero point, and $\lfloor \cdot \rceil$ denotes rounding to the nearest integer. In PTQ, $s$ and $z$ are generally determined by calibrating from a small subset of unlabeled training data. The corresponding de-quantization process is as follows:
\begin{equation}
\hat{X}_f = s\cdot (X_{Q}-z)
\enskip.
\end{equation}

The log2 quantizer $\mathcal{LQ}$ reformulates the quantization process into an exponential format, rendering it particularly suitable for scenarios characterized by long-tailed data distributions. Moreover, due to its capability for bit-shift operations, it demonstrates exceptional performance. Assuming there are only positive values in $X_f$, the computation process is as follows:
\begin{equation}
X_{Q} = \mathcal{LQ}(X_f|k) = \operatorname{clamp}(0,2^k-1,\lfloor -log_2\frac{X_f}{s} \rceil)
\enskip,
\end{equation}
\begin{equation}
s = \operatorname{max}(X_{Q})
\enskip.
\end{equation}
The corresponding de-quantization process is delineated as follows:
\begin{equation}                                               
\hat{X}_f = s \cdot 2^{-X_{Q}}
\enskip.
\end{equation}

In line with prior research on ViTs quantization \citep{RepQ-ViT}, we focus on the uniform quantizer and the log2 quantizer, considering their superior performance in real-world deployment scenarios.

\begin{figure*}[ht]
    \centering
    \includegraphics[width=\textwidth]{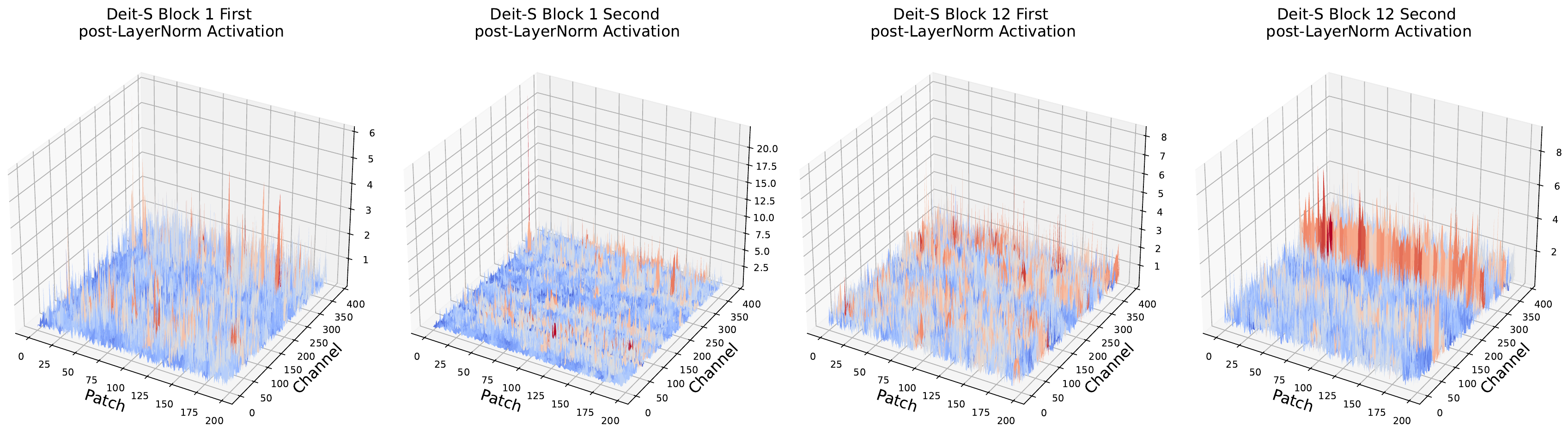}
    \caption{Visualization of the absolute value distribution for post-LayerNorm activations. We choose the DeiT-S model and select 32 images from ImageNet dataset for inference to visualize absolute value of the post-LayerNorm activations in its 1st (first) and 12th (last) blocks.}
    \label{fig:layernorm_vis}
\end{figure*}

\subsection{Per-Patch Outlier-aware Quantizer}
\label{subsec:per-patch}
For each MHA and MLP module within ViT, the input $X$ undergoes intra-sample normalization via a LayerNorm layer \citep{ba2016layer}, yielding the resultant activations $X_{ln} \in \mathbb{R}^{n \times d}$, termed as the post-LayerNorm activations. However, we observe that the distribution of $X_{ln}$ is markedly irregular, as illustrated in Figure \ref{fig:layernorm_vis}, with a minor subset of values exhibiting exceptionally large magnitudes. While earlier work \citep{RepQ-ViT} has identified an “inter-channel variation" distribution pattern for outliers in specific layers of ViTs, such as in the second post-LayerNorm activations of the 12th block shown in Figure \ref{fig:layernorm_vis}, our findings reveal that the distribution of outliers is, in fact, irregular across numerous layers, as evidenced by the additional three post-LayerNorm activations depicted in Figure \ref{fig:layernorm_vis}.

\begin{figure}[t]
    \centering
    \includegraphics[width=\columnwidth]{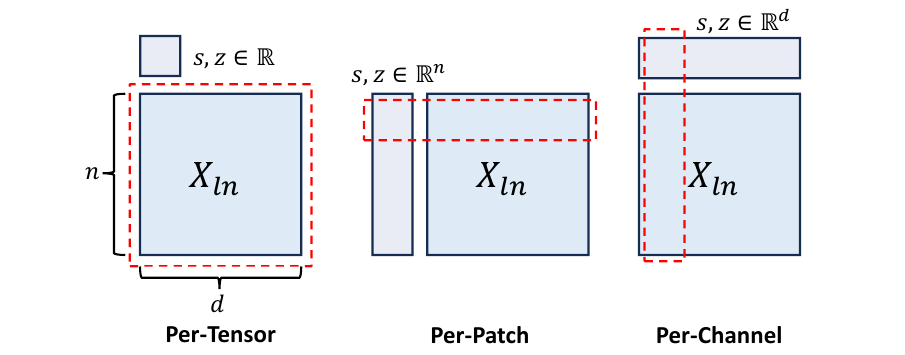}
    \caption{Comparison of different quantization granularity for post-LayerNorm activations $X_{ln}$. The red dashed boxes represent the statistical ranges of the quantization parameters.}
    \label{fig:granularity}
\end{figure}

A small part of irregularly distributed outliers in $X_{ln}$ has a direct effect on the quantization parameters including scale factor $s$, and zero point $z$ in the quantizer. Taking $s$ as an example, the presence of outliers expands the data statistical range $\operatorname{max}(X_{ln})-\operatorname{min}(X_{ln})$, resulting in a larger scale factor $s$. Consequently, this precipitates heightened rounding errors during quantization. Conversely, the adoption of a specific method to truncate the statistical range may instigate larger clipping errors during quantization. Given the constrained representational capacity at low-bit, the quantization error is further compounded. Thus, effectively attenuating the quantization error stemming from outliers within $X_{ln}$ assumes paramount importance in mitigating accuracy degradation at low-bit.

In ViTs, where the post-LayerNorm activations $X_{ln}$ serve as inputs for the QKV and FC1 Linear layers predominantly executing General Matrix Multiplication (GEMM), computed as $X_{ln}W$. This unique attribute facilitates the quantization of their inputs using either a per-tensor or per-patch granularity, as the quantization parameters can be extracted in advance during matrix computations, enabling the remaining operations to proceed using integer arithmetic. Building upon the phenomenon of irregularly distributed outliers mentioned earlier, we design per-patch granularity uniform quantizer for $X_{ln}$. Different from the per-tensor granularity commonly used in previous ViTs quantization methods, we narrow down the statistical granularity of the quantization parameters from the whole $X_{ln}$ to each patch $x_i$, and ultimately obtain $n$ different $s_i$ and $z_i$:
\begin{equation}
s_i = \frac{\operatorname{max}(x_i)-\operatorname{min}(x_i)}{2^k-1} (i = 1, ..., n)
\enskip,
\end{equation}
\begin{equation}
z_i = \lfloor \frac{-\operatorname{min}(x_i)}{s_i} \rceil
\enskip.
\end{equation}
By refining the quantization granularity to the patch dimension, patches devoid of outliers can evade the quantization errors induced by outliers, thereby effectively mitigating the accuracy loss. In addition to per-tensor and per-patch, another common quantization granularity is per-channel, and the difference is shown in Figure \ref{fig:granularity}. While per-channel quantization can also significantly alleviate the impact of outliers, the GEMM computation form of $X_{ln}W$ in the QKV and FC1 Linear layers mentioned earlier makes it difficult to extract quantization parameters when quantizing post-LayerNorm activations $X_{ln}$, resulting in deployment challenges \citep{nagel2021white}.

\begin{figure}[t]
    \centering
    \includegraphics[width=0.6\columnwidth]{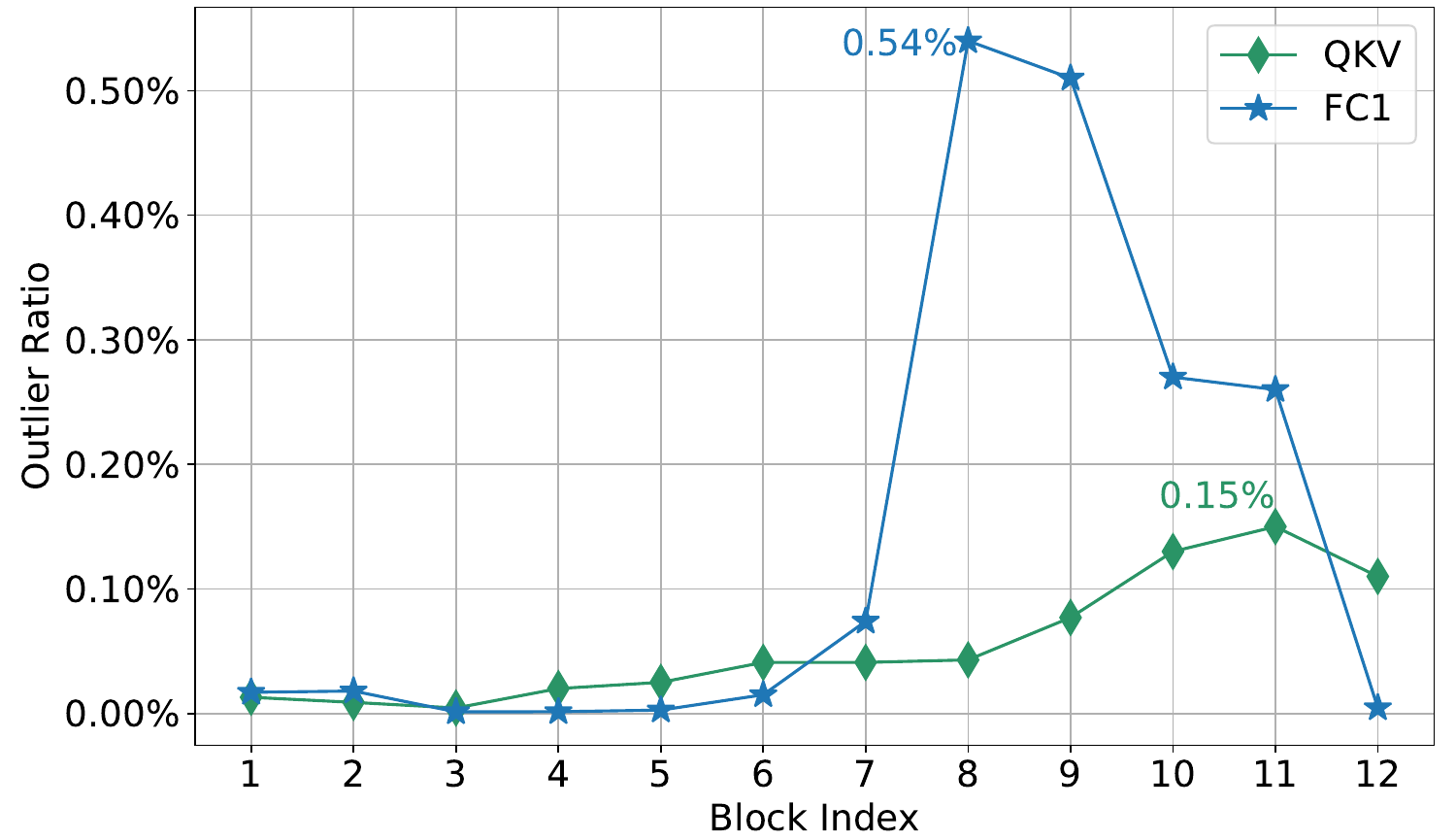}
    \caption{Ratio of outliers for input activations (post-LayerNorm activations) of QKV Linear and FC1 Liner for the Deit-S model.}
    \label{fig:outlier_ratio}
\end{figure}

Although per-patch quantization mitigates the impact of outliers by confining their effect to individual patches, outliers within a patch can still introduce quantization errors. Nonetheless, as depicted in Figure \ref{fig:outlier_ratio}, we observe that the ratio of outliers in $X_{ln}$ is exceedingly small, with most blocks having a ratio below 0.1\% and a maximum of only 0.54\%. Inspired by this finding, we propose partitioning $X_{ln}$ into two matrices, one containing outliers that exceed a pre-defined threshold $\alpha$, denoted as $M_o$, and the other comprising non-outlier values, represented as $M_{no}$. We restrict the quantization to $M_{no}$, while preserving the outlier matrix $M_o$ in full-precision. This strategy incurs negligible additional computational and storage overhead, yet it proves to be highly effective in reducing the quantization errors induced by outliers.

Combining per-patch granularity quantization and maintaining the outlier matrix $M_o$ in full-precision, we propose the Per-Patch Outlier-aware Quantizer, as shown in Figure \ref{fig:overview}. Specifically, we first obtain $M_o$ and $M_{no}$ as follows:
\begin{equation}
M_o = X_{ln} \cdot \mathbb{I}(X_{ln} \geq \alpha), M_{no}= X_{ln} - M_o
\enskip.
\end{equation}
Subsequently, each patch $m_i$ in $M_{no}$, is individually quantized and corresponding de-quantized:
\begin{equation}
m^Q_i = \mathcal{UQ}(m_i|k)
\enskip,
\end{equation}
\begin{equation}
\hat{m_i}=s_i(m^Q_i-z_i)
\enskip.
\end{equation}
The quantized matrix $M^Q_{no}$ is multiplied (GEMM) with the weight matrix $W^Q$ to obtain $Y_{no}$, while the sparse matrix $M_o$ is also multiplied (Sparse Matrix Multiplication, SpMM)) with $W^Q$ to obtain $Y_{o}$. Finally, the two results are summed to produce the final output $Y$.

\begin{figure}[t]
    \centering
    \includegraphics[width=0.9\columnwidth]{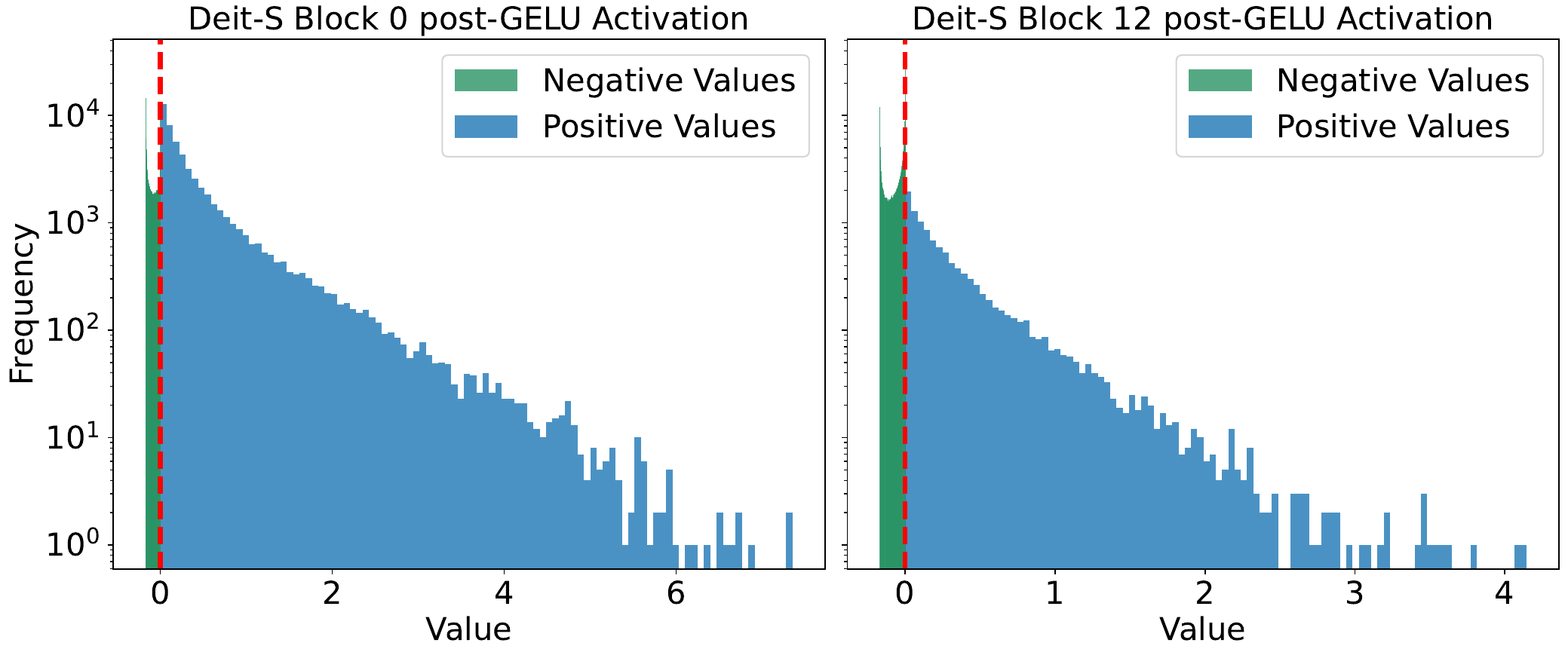}
    \caption{Visualization of the value distribution for post-GELU activations. The red dotted line indicates the dividing line between positive and negative values.}
    \label{fig:gelu_vis}
\end{figure}

\subsection{Shift-Log2 Quantizer}
\label{subsec:shift-log2}
\begin{figure}[t]
    \centering
    \includegraphics[width=0.9\columnwidth]{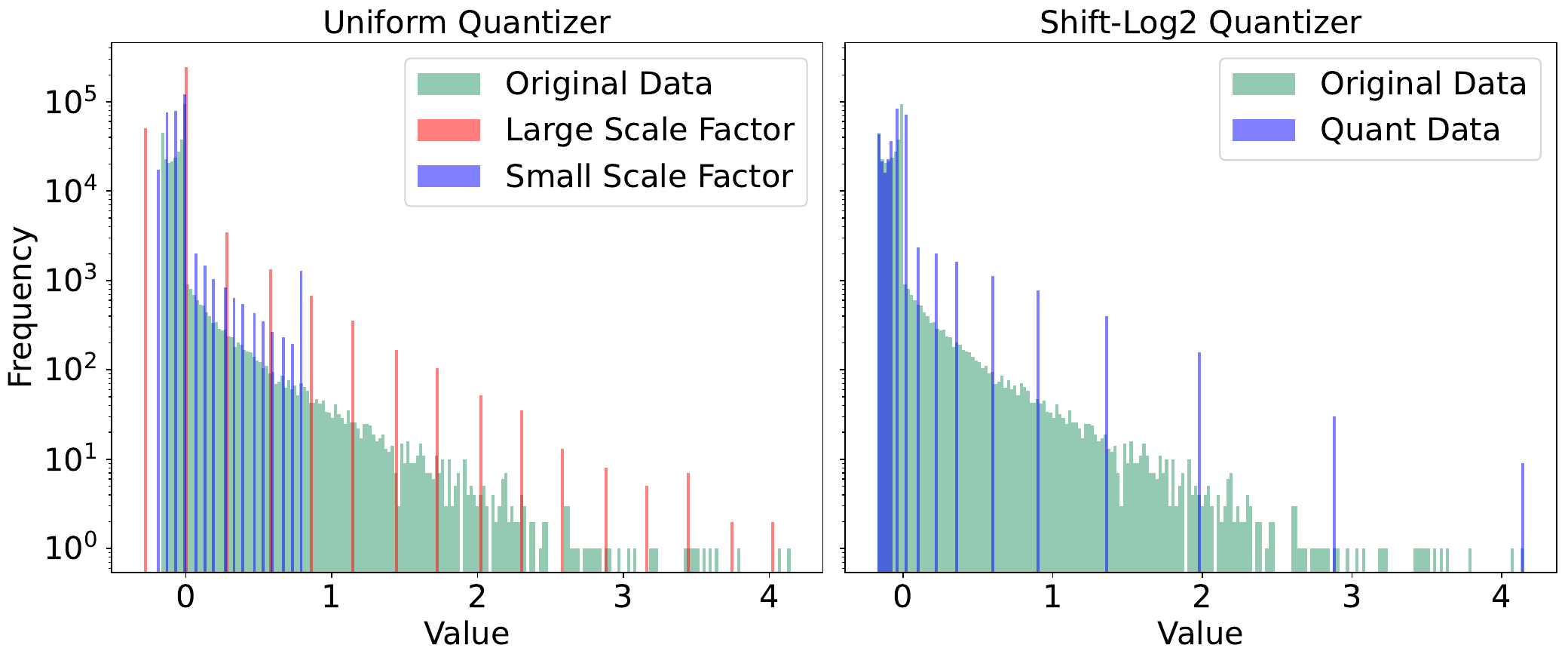}
    \caption{Comparison of the uniform quantizer and the Shift-Log2 Quantizer for post-GELU activations.}
    \label{fig:gelu_result}
\end{figure}

The GELU \citep{hendrycks2016gaussian} activation function is used inside the MLP module, defined as:
\begin{equation}
\operatorname{GELU}(x) = x\cdot P(X\leq x)
\enskip,
\end{equation}
where $P(X\leq x)$ is the cumulative distribution function (CDF) of a standard normal distribution $\mathcal{N}(0,1)$, and its output is referred to as the post-GELU activations. Figure \ref{fig:gelu_vis} illustrates that the value distribution of post-GELU activations $X_g$, is notably concentrated in its negative domain, whereas its positive domain exhibits a broader range of values. This characteristic makes it challenging to effectively quantize using a uniform quantizer. A larger scaling factor $s$ increases rounding errors in the negative region of post-GELU activations, while a smaller $s$ leads to significant clipping errors in its positive domain, as depicted in Figure \ref{fig:gelu_result}. PTQ4ViT \citep{PTQ4ViT} mitigates this issue by implementing twin uniform quantization, assigning distinct scale factors to positive and negative values, thereby minimizing quantization error through optimization. Nonetheless, this method entails considerable search optimization and deployment overhead, making it difficult to apply in practical scenarios.

Despite the non-uniform distribution across both the negative and positive regions of $X_g$, arranging the values from smallest to largest reveals a long-tailed characteristic in the data distribution. This characteristic aligns well with the log2 quantizer, which strategically allocates more quantization bins to the denser, smaller-value region, thereby minimizing rounding errors. Additionally, the log2 quantizer adeptly mitigates clipping errors in the sparser, larger-value region while ensuring comprehensive coverage across the entire data range. However, the conventional log2 quantizer assigns an equal number of quantization bins to both positive and negative values. Given the broader range of the positive value distribution, it is more efficacious to allocate a greater number of quantization bins to the positive values. Consequently, we shift all elements to be positive and subsequently employ the positive-only log2 quantizer, as delineated in Section 3.1, for quantization.

Building on these insights and maximizing the utilization of quantization bins, we introduce the Shift-Log2 Quantizer. We start by shifting all elements of $X_g$ to ensure they are greater than 0, becoming $X_g'$:
\begin{equation}
X_g'=X_g-\operatorname{min}(X_g)+\epsilon
\enskip,
\end{equation}
where $\epsilon$ is a very small positive constant. The shifted values are then quantized using the log2 quantizer:
\begin{equation}
X_{g}^Q = \mathcal{LQ}(X_g'|k)
\enskip,
\end{equation}
and then, it is shifted back to its original value during de-quantization:
\begin{equation}
\hat{X}_g = s \cdot 2^{-X_{g}^Q} + \operatorname{min}(X_g) - \epsilon
\enskip.
\end{equation}
By employing this quantizer, we effectively preserve the intrinsic characteristics of the activations, significantly reduce quantization error, and avoid excessive computational overhead. Compared to the uniform quantizer, the Shift-Log2 Quantizer better preserves the distribution of the original values, as demonstrated in Figure \ref{fig:gelu_result}.

\subsection{Attention-score enhanced Module-wise Optimization}
\label{subsec:attention-score}
To further mitigate the accuracy loss of ViTs under low-bit quantization, we adopt an error reconstruction method to optimize the parameters of the weights and activation quantizers. Different from previous work \citep{BRECQ}, we transform the reconstruction granularity from block-wise to module-wise. In this context, ``block-wise" refers to tuning aimed at an entire transformer block, whereas ``module-wise" is optimized for MHA and MLP modules respectively. With fine-grained tuning, we can better minimize the quantization error.

For the weight quantizer, we adopt the method in AdaRound \citep{nagel2020up} to introduce an optimizable matrix $V$ to decide whether rounding is up or down, and the specific quantization process is as follows:
\begin{equation}
W_{Q}=\mathcal{UQ}(W_f|k) = \operatorname{clamp}(0,2^{k}-1,\lfloor \frac{W_f}{s} \rfloor + h(V))
\enskip,
\end{equation}
$\lfloor . \rfloor$ is rounded down, where define  rectified sigmoid \citep{louizos2018learning} as $h(V)$:
\begin{equation}
h(V) = \operatorname{clamp}(0,1,\sigma(V)(\zeta-\gamma)+\gamma)
\enskip,
\end{equation}
the $\sigma$ is a sigmoid function; $\zeta$ and $\gamma$ fixed at 1.1 and -0.1, respectively. Our main optimization is $V$ in weight quantizer. Regarding the activation quantizer, we mainly optimize the scaling factor $s$, which we define as $s_a$ to distinguish it.

Specifically, in quantizing the MLP module, our optimization objective is defined as:
\begin{equation}
\underset{V,s_a} {{\arg\min} }\mathcal{L}_{mlp}
\enskip,
\end{equation}
where $\mathcal{L}_{mlp}$:
\begin{equation}
\mathcal{L}_{mlp}=\mathcal{L}_o+\lambda\mathcal{L}_{round}
\enskip,
\end{equation}
$\lambda$ is a balance coefficient, $\mathcal{L}_o$ is defined as the MSE between the output $X_{mlp}$ and $\hat{X}_{mlp}$ of the module before and after quantization:
\begin{equation}
\mathcal{L}_o = ||X_{mlp}-\hat{X}_{mlp}||^2_2
\enskip.
\end{equation}
And $\mathcal{L}_{round}$ is defined as the rounding error of the weights:
\begin{equation}
\mathcal{L}_{round} = \underset{i,j}\sum1-|2h(V_{i,j})-1|^\beta
\enskip,
\end{equation}
$\beta$ is a continuous linear annealing parameter ranging from 10 to 2. Compared to the MLP module, in quantizing the MHA module, since attention score $AS$ is one of the critical attributes, $\mathcal{L}_{mha}$ is defined as:
\begin{equation}
\mathcal{L}_{mha} = \mathcal{L}_{as}+\mathcal{L}_o+\lambda\mathcal{L}_{round}
\enskip,
\end{equation}
where $\mathcal{L}_{as}$ is:
\begin{equation}
\mathcal{L}_{as} = \operatorname{KL}(AS,\hat{AS})
\enskip,
\end{equation}
KL refers to the Kullback-Leibler divergence. We denote the above whole process Attentin-score enhanced Module-wise Optimization, which ensures that the critical characteristics of the attention scores within the MHA module are preserved as accurately as possible during the quantization process. This optimization effectively minimizes quantization error, particularly in scenarios involving low-bit quantization.

\subsection{Quantization Procedure}
\label{subsec:procedure}
The quantization procedure of our ADFQ-ViT framework is essentially divided into two key stages. Initially, the procedure commences by substituting specific modules in the original model with corresponding quantization modules. These modules are equipped with the requisite functions for both weight and activation quantizers, involving the initialization of appropriate parameters, as depicted in the middle part of Figure \ref{fig:overview}. Subsequently, the process involves selecting a certain amount of unlabeled data and proceeding through a specified number of iteration steps to optimize the parameters of the quantizers, as illustrated in the right section of Figure \ref{fig:overview}. Upon the completion of these stages, the final quantized model is prepared for deployment.

%% file: tex/experiments.tex
\section{Experiments}
\subsection{Experimental Setup}
\textbf{Tasks, Models And Datasets.} We evaluate ADFQ-ViT across a variety of vision tasks, including image classification, object detection, and instance segmentation. For image classification, we select three distinct models: ViT \citep{vit}, DeiT \citep{deit}, and Swin Transformer \citep{swin}, and assess their Top-1 accuracy on the ImageNet \citep{krizhevsky2012imagenet} dataset. For object detection and instance segmentation, we utilize Mask R-CNN \citep{he2017mask} and Cascade Mask R-CNN \citep{cai2018cascade}, with Swin Transformer as the backbone, and evaluate their performance using the $AP^{box}$ and $AP^{mask}$ metrics on the COCO \citep{lin2014microsoft} dataset. Additionally, we evaluate the zero-shot instance segmentation performance of the SAM on the COCO dataset, using $AP^{mask}$ as the metric.

\textbf{Implementation Details.} Given that our ADFQ-ViT framework falls under the PTQ category, obtaining the corresponding pre-trained models before quantization is essential. For image classification task, all pre-trained models are sourced from the Timm library\footnote{https://github.com/rwightman/pytorch-image-models}. For object detection and instance segmentation, the pre-trained models come from the official Swin Transformer implementation based on the mm-detection library\footnote{https://github.com/SwinTransformer/Swin-Transformer-Object-Detection}. The SAM is obtained from the Prompt-Segment-Anything library\footnote{https://github.com/RockeyCoss/Prompt-Segment-Anything}. Regarding the selection of calibration samples for the quantization process, we select 1,024 samples from the ImageNet training set as the calibration set for the image classification task. For object detection and instance segmentation (including SAM), we select 1 sample from the COCO dataset as the calibration set. Consistent with previous work \citep{PTQ4ViT}, we quantize the weights and input activations of all matrix multiplication layers, but not the LayerNorm and Softmax layers. All weight quantizers are per-channel uniform quantizers. For post-LayerNorm activations, a Per-Patch Outlier-aware Quantizer is applied, while post-GELU activations use the Shift-Log2 Quantizer. Post-Softmax activations use the log2 quantizer, and all other activations are quantized using per-tensor uniform quantizers. During the Attention-score enhanced Module-wise Optimization, the learning rate for the weight quantizer parameters is set to 3e-3, and for the activation quantizer parameters, it is configured to 4e-5. Both parameters are optimized using the Adam optimizer \citep{kingma2014adam}, and the entire process runs for 3000 iterations. The experimental implementations are conducted using PyTorch \citep{paszke2019pytorch} on an NVIDIA 3090 GPU.

\begin{table}[htbp]
\begin{adjustbox}{width=\columnwidth}
\centering
\begin{tabular}{cccccccccc}
\specialrule{1.5pt}{0pt}{0pt}
\addlinespace
\textbf{Method} & \textbf{W-bit} & \textbf{A-bit} & \textbf{ViT-S} & \textbf{ViT-B} & \textbf{DeiT-T} & \textbf{DeiT-S} & \textbf{DeiT-B} & \textbf{Swin-S} & \textbf{Swin-B} \\
\midrule
Full-Precision & 32 & 32 & 81.39 & 84.54 & 72.21 & 79.85 & 81.80 & 83.23 & 85.27 \\
\midrule
\addlinespace
BRECQ \citep{BRECQ} & 4 & 4 & 12.36 & 9.68 & 55.63 & 63.73 & 72.31 & 72.74 & 58.24 \\
QDrop \citep{Qdrop}& 4 & 4 & 21.24 & 47.30 & 61.93 & 68.27 & 72.60 & 79.58 & 80.93 \\
PD-Quant \citep{pdquant}& 4 & 4 & 1.51 & 32.45 & 62.46 & 71.21 & 73.76 & 79.87 & 81.12 \\
PTQ4ViT \citep{PTQ4ViT} & 4 & 4 & 42.57 & 30.69 & 36.96 & 34.08 & 64.39 & 76.09 & 74.02 \\
APQ-ViT \citep{ding2022towards} & 4 & 4 & 47.95 & 41.41 & 47.94 & 43.55 & 67.48 & 77.15 & 76.48 \\
RepQ-ViT \citep{RepQ-ViT}& 4 & 4 & 65.05 & 68.48 & 57.43 & 69.03 & 75.61 & 79.45 & 78.32 \\
\textbf{ADFQ-ViT (Ours)} & 4 & 4 & \textbf{72.14} & \textbf{78.71} & \textbf{62.91} & \textbf{75.06} & \textbf{78.75} & \textbf{80.63} & \textbf{82.33} \\
\midrule
\addlinespace
BRECQ \citep{BRECQ}& 6 & 6 & 54.51 & 68.33 & 70.28 & 78.46 & 80.85 & 82.02 & 83.94 \\
QDrop \citep{Qdrop} & 6 & 6 & 70.25 & 75.76 & 70.64 & 77.95 & 80.87 & 82.60 & 84.33 \\
PD-Quant\citep{pdquant} & 6 & 6 & 70.84 & 75.82 & 70.49 & 78.40 & 80.52 & 82.51 & 84.32 \\
PTQ4ViT \citep{PTQ4ViT} & 6 & 6 & 78.63 & 81.65 & 69.68 & 76.28 & 80.25 & 82.38 & 84.01 \\
APQ-ViT \citep{ding2022towards} & 6 & 6 & 79.10 & 82.21 & 70.49 & 77.76 & 80.42 & 82.67 & 84.18 \\
RepQ-ViT\citep{RepQ-ViT} & 6 & 6 & 80.43 & 83.62 & 70.26 & 78.90 & 81.27 & 82.79 & 84.57 \\
\textbf{ADFQ-ViT (Ours)} & 6 & 6 & \textbf{80.54} & \textbf{83.92} & \textbf{70.96} & \textbf{79.04} & \textbf{81.53} & \textbf{82.81} & \textbf{84.82} \\
\bottomrule
\end{tabular}
\end{adjustbox}
\caption{Comparison of accuracy on image classification task. The “W-bit” and “A-bit” represent the quantization bit-width of weights and activations, respectively, and the evaluation metric is the Top-1 accuracy on ImageNet dataset; bolding indicates the optimal result for corresponding settings.}
\label{tab:classification}
\end{table}

\begin{table}[h]
  \centering
  \begin{adjustbox}{width=\textwidth}
  \begin{tabular}{ccccccccccc}
    \specialrule{1.5pt}{0pt}{0pt}
    \multirow{3}{*}{\textbf{Method}} & \multirow{3}{*}{\textbf{W-bit}} & \multirow{3}{*}{\textbf{A-bit}}  & \multicolumn{4}{c}{\textbf{Mask R-CNN}} & \multicolumn{4}{c}{\textbf{Cascade Mask R-CNN}}\\
    \cmidrule(lr){4-7} \cmidrule(lr){8-11} 
    & & & \multicolumn{2}{c}{\textbf{w. Swin-T}} & \multicolumn{2}{c}{\textbf{w. Swin-S}} & \multicolumn{2}{c}{\textbf{w. Swin-T}} & \multicolumn{2}{c}{\textbf{w. Swin-S}} \\
    & & & \textbf{$\operatorname{AP}^{box}$} & \textbf{$\operatorname{AP}^{mask}$} & \textbf{$\operatorname{AP}^{box}$} & \textbf{$\operatorname{AP}^{mask}$} & \textbf{$\operatorname{AP}^{box}$} & \textbf{$\operatorname{AP}^{mask}$} & \textbf{$\operatorname{AP}^{box}$} & \textbf{$\operatorname{AP}^{mask}$} \\
    \midrule
    Full-Precision & 32 & 32 & 46.0 & 41.6 & 48.5 & 43.3 & 50.4 & 43.7 & 51.9 & 45.0 \\
    \midrule
    \addlinespace
    PTQ4ViT \citep{PTQ4ViT}& 4 & 4 & 6.9 & 7.0 & 26.7 & 26.6 & 14.7 & 13.5 & 0.5 & 0.5 \\
    APQ-ViT \citep{ding2022towards}& 4 & 4 & 23.7 & 22.6 & \textbf{44.7} & 40.1 & 27.2 & 24.4 & 47.7 & 41.1 \\
    RepQ-ViT \citep{RepQ-ViT}& 4 & 4 & 36.1 & 36.0 & 42.7 & 40.1 & 47.0 & 41.4 & 49.3 & 43.1 \\
    \textbf{ADFQ-ViT (ours)} & 4 & 4 & \textbf{36.9} & \textbf{36.4} & 43.9 & \textbf{40.9} & \textbf{48.3} & \textbf{42.5} & \textbf{49.5} & \textbf{43.2} \\
    \midrule
    \addlinespace
    PTQ4ViT \citep{PTQ4ViT}& 6 & 6 & 5.8 & 6.8 & 6.5 & 6.6 & 14.7 & 13.6 & 12.5 & 10.8 \\
    APQ-ViT \citep{ding2022towards}& 6 & 6 & 45.4 & 41.2 & 47.9 & 42.9 & 48.6 & 42.5 & 50.5 & 43.9 \\
    RepQ-ViT \citep{RepQ-ViT}& 6 & 6 & 45.1 & 41.2 & 47.8 & \textbf{43.0} & 50.0 & 43.5 & 51.4 & 44.6 \\
    \textbf{ADFQ-ViT (ours)} & 6 & 6 & \textbf{45.6} & \textbf{41.4} & \textbf{48.0} & \textbf{43.0} & \textbf{50.2} & \textbf{43.6} & \textbf{51.8} & \textbf{44.9} \\
    \bottomrule
  \end{tabular}
  \end{adjustbox}
  \caption{Comparison of accuracy on object detection and instance segmentation tasks. $AP^{box}$ is the box average accuracy for object detection, and $AP^{mask}$ is the mask average accuracy for instance segmentation.}
  \label{tab:detection}
\end{table}

\subsection{Comparison with Baselines} To demonstrate the effectiveness of our ADFQ-ViT framework, we select six diverse and representative quantization methods as baselines for comparison over three computer vision tasks. These baselines include three CNNs quantization methods: BRECQ, QDrop, and PD-Quant, as well as three ViTs quantization methods: APQ-ViT, PTQ4ViT, and RepQ-ViT. To ensure a fair comparison, we use the baseline accuracy reported in \citep{zhong2023s} and \citep{li2024repquant}, as their experiments are conducted under conditions comparable to ours in terms of tasks, models, and datasets.

\textbf{Image Classification.} First, we evaluate our method and the six baseline methods on the ImageNet dataset, comparing the Top-1 classification accuracy after 4-bit and 6-bit quantization across seven different models with varying architectures and parameter sizes (ViT, DeiT and denote Swin Transformer by Swin; T for Tiny parameter size, S for Small, B for Base). The experimental results are shown in Table \ref{tab:classification}. At 4-bit quantization, our method shows a significant accuracy improvement over the best corresponding baseline across all models, with an average increase of 3.82\%. Notably, previous methods suffer severe accuracy drops when quantizing ViT-S and ViT-B, whereas our method improves accuracy by 7.09\% and 10.23\% compared to RepQ-ViT. Furthermore, at 6-bit quantization, our method achieves a consistent accuracy improvement over the baselines. Compared to the full-precision models, our method results in an average accuracy drop of only 0.67\%, which is very close to lossless performance. Meanwhile, we observe that on DeiT and Swin Transformer, the CNNs quantization methods achieve performance comparable to that of the previous ViTs quantization, which is due to the fact that all of these CNNs quantization use reconstruction optimization to tune the quantization parameters, illustrating the effectiveness of optimizing quantization parameters. However, the performance of these CNNs quantization methods falls short compared to ViTs quantization methods on ViT models, primarily because they are designed for CNNs, and ViT exhibits more distinct distribution characteristics compared to DeiT and Swin Transformer. Our method considers the distributional properties of ViTs and combines the advantages of reconstruction optimization to achieve optimal performance.

\textbf{Object Detection And Instance Segmentation.} Next, we evaluate our method and three ViTs quantization methods on object detection and instance segmentation tasks using Mask R-CNN and Cascade Mask R-CNN on the COCO dataset. We select two different models, Swin-T and Swin-S, as backbones for evaluation. The experimental results at 4-bit and 6-bit quantization are shown in Table \ref{tab:detection}. Compared to image classification, object detection and instance segmentation tasks are more challenging, which results in a significant accuracy drop for most methods when quantized to 4-bit. Despite these challenges, our method consistently achieves improved accuracy compared to the baselines in most models at 4-bit quantization. Remarkably, our method surpasses the baselines on Cascade Mask R-CNN, both employing Swin-T or Swin-S as the backbone and achieving improvements of 1.3 in $AP^{box}$ and 1.1 in $AP^{mask}$, especially for Swin-T backbone. While APQ-ViT achieves a 0.8 improvement in $AP^{box}$ when using Swin-S as the backbone for Mask R-CNN, this method lacks stability compared to ours, experiencing severe accuracy loss in other models applied with different backbones. At 6-bit quantization, our method achieves the best performance compared to the baselines across all models. Moreover, compared to the original full-precision models, our method experiences an average accuracy drop of only 0.3 in $AP^{box}$, and 0.17 in $AP^{mask}$. Particularly, on the Cascade Mask R-CNN with Swin-S model, which achieves the highest accuracy, our method only experiences a 0.1 drop in both $AP^{box}$ and $AP^{mask}$. This makes deploying quantized models in real-world environments feasible, enhancing efficiency and feasibility in real-time and resource-constrained scenarios.

\begin{table}[t]
  \centering
  \begin{adjustbox}{width=0.7\textwidth}
  \begin{tabular}{ccccc}
    \specialrule{1.5pt}{0pt}{0pt}
   \textbf{SAM Model} & \textbf{Method} & \textbf{W-bit} & \textbf{A-bit} & \textbf{$AP^{mask}$} \\
    \midrule
    \addlinespace
    \multirow{7}{*}{\textbf{ViT-B}} & Full-Precision & 32 & 32 & 42.5 \\
    \cmidrule(lr){2-5}
    & PTQ4-ViT & 4 & 4 & 8.4 \\
    & RepQ-ViT & 4 & 4 & 37.6 \\
    & \textbf{ADFQ-ViT (ours)} & 4 & 4 & \textbf{38.6} \\
    \cmidrule(lr){2-5}
    & PTQ4-ViT & 6 & 6 & 8.9 \\
    & RepQ-ViT & 6 & 6 & 42.0 \\
    & \textbf{ADFQ-ViT (ours)}& 6 & 6 & \textbf{42.2} \\
    \midrule
    \addlinespace
    \multirow{7}{*}{\textbf{ViT-L}} & Full-Precision & 32 & 32 & 46.3 \\
    \cmidrule(lr){2-5}
    & PTQ4ViT & 4 & 4 & 11.3 \\
    & RepQ-ViT & 4 & 4 & 40.7 \\
    & \textbf{ADFQ-ViT (ours)} & 4 & 4 & \textbf{44.7} \\
    \cmidrule(lr){2-5}
    & PTQ4ViT & 6 & 6 & 13.5 \\
    & RepQ-ViT & 6 & 6 & 45.8 \\
    & \textbf{ADFQ-ViT (ours)} & 6 & 6 & \textbf{46.2} \\
    \bottomrule
  \end{tabular}
  \end{adjustbox}
  \caption{Comparison of accuracy on zero-shot instance segmentation of
SAM model on COCO dataset. “SAM Model” here represents the image encoder used. H-Deformable-DETR with Swin-L is used as the object detector and the model is prompted using the boxes. $AP^{mask}$ is the mask
average accuracy for instance segmentation.}
  \label{tab:sam}
\end{table}

\begin{figure*}[t]
    \centering
    \includegraphics[width=\textwidth]{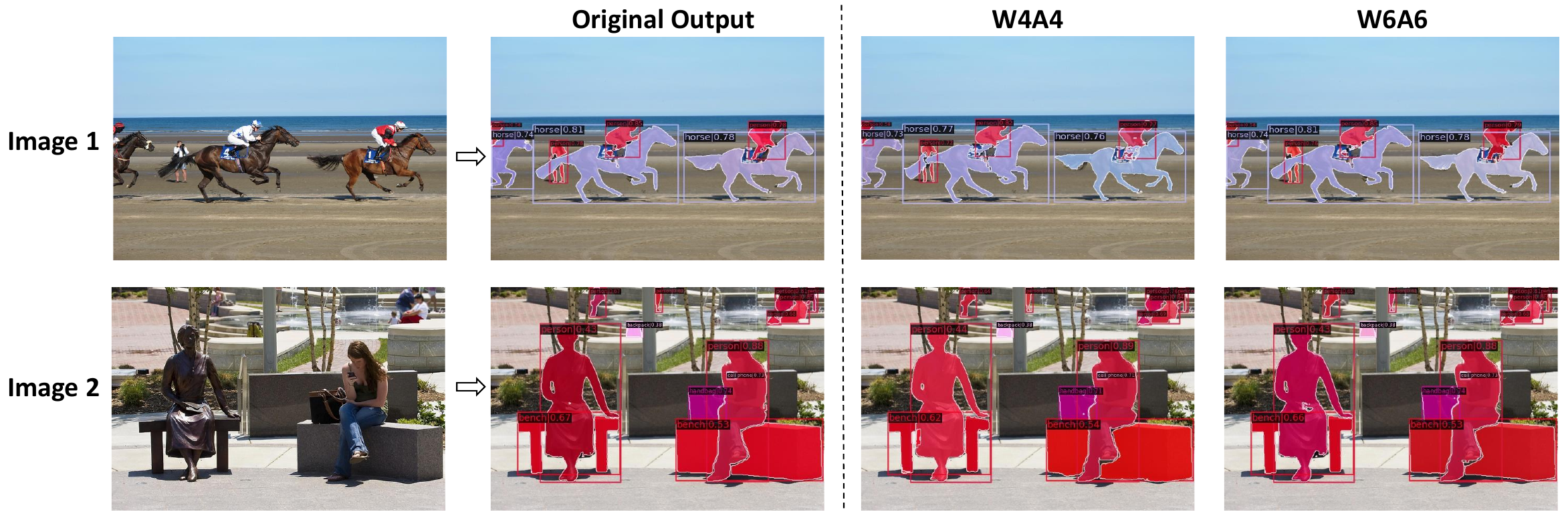}
    \caption{Visualization of zero-shot instance segmentation results. We select two images from the COCO dataset to compare the segmentation results of the original full-precision model with those of the quantized model. The SAM model uses ViT-B as the image encoder, while H-Deformable-DETR with Swin-L is used as the object detector, and the model is prompted using the boxes. “W4A4" indicates that both weights and activations are quantized to 4-bit, and “W6A6" indicates that both are quantized to 6-bit.}
    \label{fig:sam_vis}
\end{figure*}

\textbf{Segment Anything Model.} Finally, to validate the broad applicability of our method, we conduct experiments on SAM, which has garnered significant attention for its exceptional zero-shot instance segmentation performance. The SAM model consists of three components: the image encoder, mask decoder, and prompt decoder. We focus on quantizing the image encoder of SAM, keeping the other components at full-precision. Concretely, we evaluate $AP^{mask}$ of SAM on the COCO dataset under 4-bit and 6-bit quantization, using ViT-B and ViT-L as the image encoder, as shown in Table \ref{tab:sam}. At 4-bit quantization, our method enhances $AP^{mask}$ by 1.0 for SAM with ViT-B and by 4.0 for SAM with ViT-L compared to RepQ-ViT. At 6-bit quantization, our method demonstrates improvements of 0.2 and 0.4, respectively, with only a 0.3 and 0.1 decrease compared to the full-precision model. Furthermore, as illustrated in Figure \ref{fig:sam_vis}, we visualize the segmentation results of SAM after 4-bit and 6-bit quantization. The segmentation quality remains consistent with the original full-precision model, even at 4-bit. These results indicate that our method sustains excellent performance on SAM and highlights its potential applicability to complex models using ViTs as the foundational structure.

\begin{table}[t]
  \centering
  \begin{adjustbox}{width=0.6\textwidth}
  \begin{tabular}{ccccc}
    \specialrule{1.5pt}{0pt}{0pt}
   \textbf{Model} & \textbf{POQ} & \textbf{SLQ} & \textbf{AMO} & \textbf{Top-1 Accuracy} \\
    \midrule
    \addlinespace
    \multirow{8}{*}{\textbf{DeiT-S}} & \ding{55} & \ding{55} & \ding{55} & 24.79 \\
    \cmidrule(lr){2-5} 
    & \ding{51} & \ding{55} & \ding{55} & 49.86 \\
    & \ding{55} & \ding{51} & \ding{55} & 41.16 \\
    & \ding{55} & \ding{55} & \ding{51} & 61.17 \\
    & \ding{51} & \ding{51} & \ding{55} & 70.37 \\
    & \ding{51} & \ding{55} & \ding{51} & 61.23 \\
    & \ding{55} & \ding{51} & \ding{51} & 70.83 \\
    \cmidrule(lr){2-5} 
    & \ding{51} & \ding{51} & \ding{51} & \textbf{75.06} \\
    \bottomrule
  \end{tabular}
  \end{adjustbox}
  \caption{Ablations study of three main components of the proposed ADFQ-ViT framework on 4-bit quantized DeiT-S. (POQ: Per-Patch Outlier-aware Quantizer, SLQ: Shift-Log2 Quantizer, AMO: Attention-score enhanced Module-wise Optimization)}
  \label{tab:ablation}
\end{table}

\subsection{Ablation Study}
\textbf{Effect of Each Component.} Our ADFQ-ViT consists of three core components: Per-Patch Outlier-aware Quantizer, Shift-Log2 Quantizer, and Attention-score enhanced Module-wise Optimization. To demonstrate the effectiveness of each component, we select DeiT-S model for ablation experiments in the image classification task to analyze the impact of each component. The results, shown in Table \ref{tab:ablation}, highlight the contributions of each component. With all components disabled, the accuracy drops significantly to 24.79\%, reflecting the challenges of low-bit quantization for ViTs. Enabling a single component, the Attention-score enhanced Module-wise Optimization shows the most significant improvement, achieving an accuracy of 61.17\%. This underscores the effectiveness of adjusting quantizer parameters, yet there remains a considerable loss, indicating the need for optimizations tailored to the unique distribution characteristics of ViTs. When enabling two components, the combination of Per-Patch Outlier-aware Quantizer and Shift-Log2 Quantizer achieves an accuracy of 70.37\%. This demonstrates that focusing on addressing the issue of the unique distributions of post-LayerNorm and post-GELU activations can effectively reduce the accuracy loss under low-bit quantization for ViTs. Finally, with all three components enabled, the accuracy reaches 75.06\%, confirming the effectiveness of ADFQ-ViT in designing suitable quantizers considering the distributional characteristics of ViTs while tuning the quantizers parameters.

\begin{table}[t]
  \centering
  \begin{adjustbox}{width=0.6\textwidth}
  \begin{tabular}{ccc}
    \specialrule{1.5pt}{0pt}{0pt}
   \textbf{Model} & \textbf{Method} & \textbf{Top-1 Accuracy} \\
    \midrule
    \addlinespace
    \multirow{2}{*}{\textbf{ViT-B}} & w/o Outlier-Aware & 76.75 \\
    & w/ Outlier-Aware & \textbf{78.71}  \\
    \midrule
    \multirow{2}{*}{\textbf{DeiT-S}} & w/o Outlier-Aware & 73.52 \\
    & w/ Outlier-Aware & \textbf{75.06}  \\
    \bottomrule
  \end{tabular}
  \end{adjustbox}
  \caption{Ablation study of Per-Patch Outlier-aware Quantizer on 4-bit quantized ViT-B and DeiT-S.}
  \label{tab:ablation_outlier}
\end{table}



\begin{figure}[t]
\centering
  \subfloat{\includegraphics[width=0.49\textwidth]{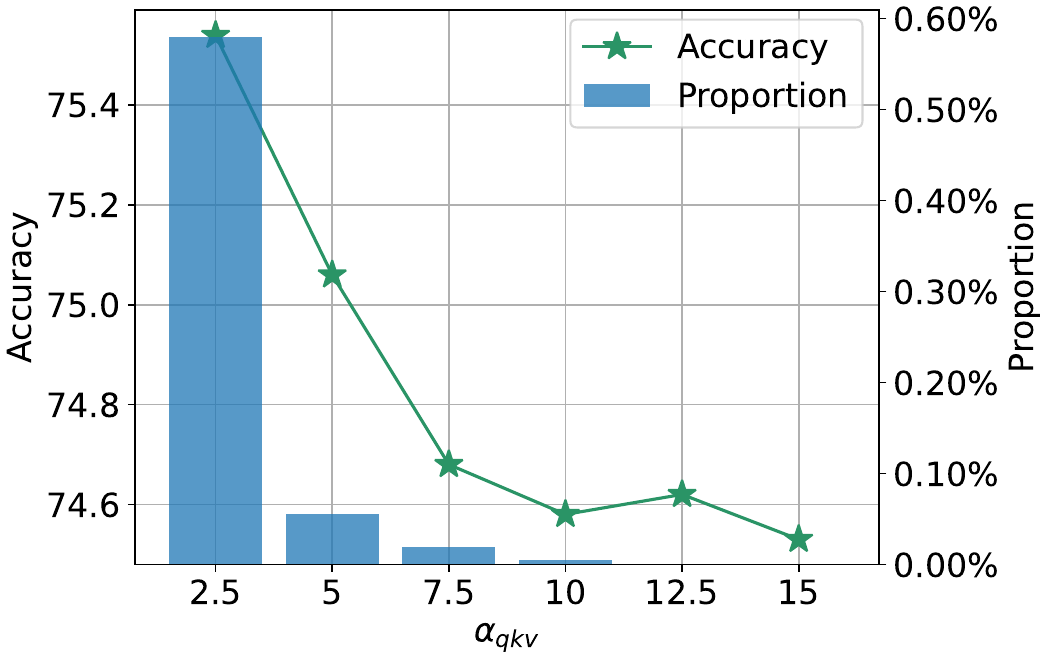}} 
  \subfloat{\includegraphics[width=0.49\textwidth]{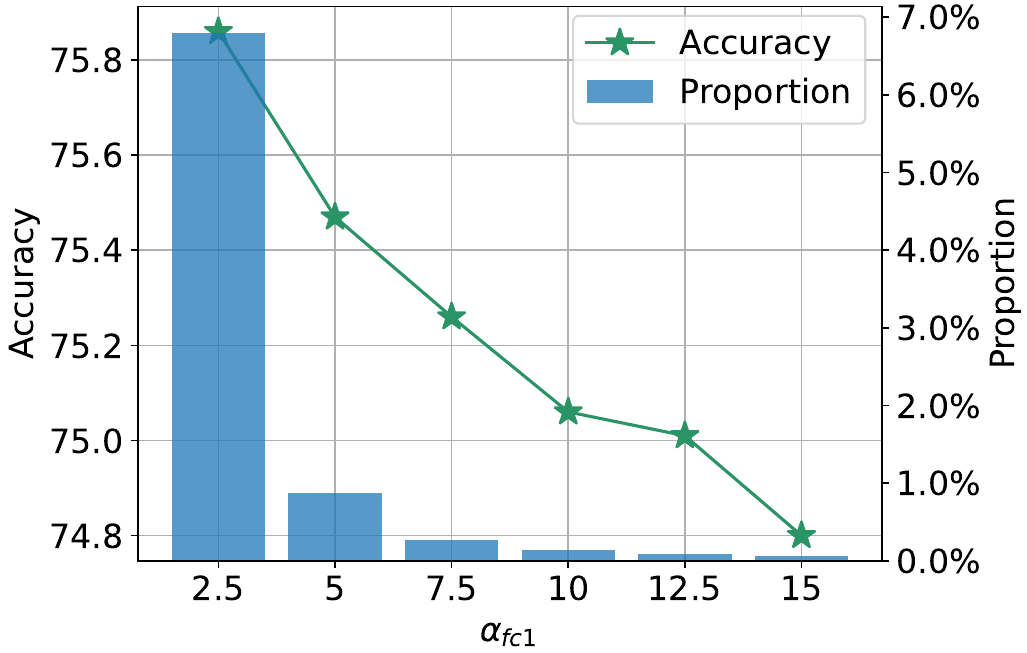}}
\caption{Ablation study of different threshold $\alpha$ for QKV and FC1 Linear layer on 4-bit quantized DeiT-S.}
\label{fig:outlier4akv_fc1}
\end{figure}

\textbf{Effect of Outlier-aware.} We conduct an in-depth analysis of the Per-Patch Outlier-aware Quantizer in ADFQ-ViT, focusing primarily on the effect of preserving outliers at their original full-precision. First, we select the ViT-B and DeiT-S and evaluate the accuracy of 4-bit quantization with and without outlier-aware enabled for the image classification task. The results are shown in Table \ref{tab:ablation_outlier}. Based on the results, enabling outlier-aware quantization improves accuracy by 1.96\% on ViT-B and 1.54\% on DeiT-S, demonstrating its effectiveness. Since the “outlier-aware" mechanism is influenced by a pre-defined threshold $\alpha$, we analyze the impact of different $\alpha$ values on the proportion of outliers and the accuracy of DeiT-S for image classification at 4-bit quantization. This analysis applies to both the QKV Linear and FC1 Linear layers, as shown in Figure \ref{fig:outlier4akv_fc1}. When adjusting $\alpha$ for one layer, the other layer’s $\alpha$ remains constant at 5 for QKV Linear and 10 for FC1 Linear. As $\alpha_{qkv}$ increases, the proportion of outliers consistently decreases; specifically, when $\alpha_{qkv}$ is set to 5, the outlier proportion is already below 0.05\%. Similarly, for $\alpha_{fc1}$ set to 10, the outlier proportion approaches 0.1\%. A smaller proportion of outliers enhances efficiency. However, accuracy decreases with increasing $\alpha_{qkv}$ and $\alpha_{fc1}$ due to the expanded quantization range. Therefore, to maintain optimal quantization accuracy, we set $\alpha_{qkv}$ to 5 and $\alpha_{fc1}$ to 10 after considering trade-offs.

\subsection{Efficiency Analysis}
\begin{figure}[t]
    \centering
    \includegraphics[width=0.5\columnwidth]{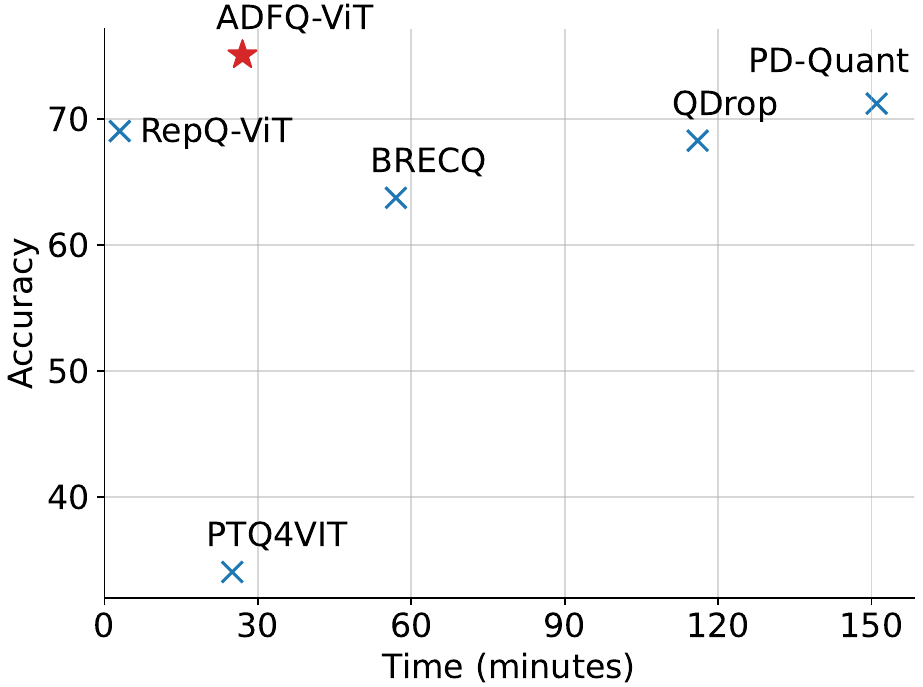}
    \caption{Comparison of quantization runtime and accuracy on 4-bit quantized DeiT-S.}
    \label{fig:quantization_runtime}
\end{figure}

Our ADFQ-ViT framework encompasses quantization parameter calibration and reconstruction optimization processes, which introduce additional overhead to the quantization process prior to actual inference. To evaluate the efficiency of ADFQ-ViT, we compare the quantization runtime of existing quantization methods for DeiT-S at 4-bit, shown in Figure \ref{fig:quantization_runtime}. The quantization runtime of ADFQ-ViT is greater than RepQ-ViT, primarily because RepQ-ViT does not involve the reconstruction process of quantization parameters, while ADFQ-ViT applies Attention-score enhanced Module-wise Optimization to update the assigned quantization parameters further to achieve minimal quantization error. In contrast, ADFQ-ViT has a much smaller quantization runtime compared to several other methods which also perform quantization parameter optimization. This efficiency is primarily due to our shift to module-wise granularity and the inclusion of the key attribute, attention score, as an optimization loss. Overall, ADFQ-ViT achieves the highest accuracy among quantized DeiT-S model with an acceptable quantization runtime, underscoring its effectiveness and efficiency in practical scenarios.

%% file: tex/conclusion.tex
\section{Conclusion}
In this paper, we propose ADFQ-ViT to address the issue of significant accuracy loss of ViT quantization at low-bit. More specifically, the irregular distribution of outliers of post-LayerNorm activations and the non-uniform distribution of positive and negative regions of post-GELU activations cause the quantization error for conventional quantizers. We apply Per-Patch Outlier-aware Quantizer, Shift-Log2 Quantizer, and Attenion-score enhanced Module-wise Optimization to tackle the challenge of activation distribution, and an extensive array of experimental results demonstrate the efficacy of this quantization framework. In the future, we aim to further explore the quantization performance of ViTs under more specific tasks, such as real-time object detection, or large multimodal models.